\documentclass{amia}
\usepackage{graphicx}
\usepackage{xcolor}
\usepackage{bm}
\usepackage{bbm}
\usepackage{tabularx,colortbl}
\usepackage{hyperref}
\usepackage[labelfont=bf]{caption}
\usepackage[nomove]{cite}

\begin{document}

\title{Learning Predictive and Interpretable Timeseries Summaries from ICU Data}

\author{Nari Johnson, MSc, Sonali Parbhoo, PhD, Andrew S Ross, PhD, Finale Doshi-Velez, PhD }

\institutes{
    School of Engineering and Applied Sciences, Harvard University, Cambridge, MA\\
}

\maketitle

\noindent{\bf Abstract}

\textit{Machine learning models that utilize patient data across time (rather than just the most recent measurements) have increased performance for many risk stratification tasks in the intensive care unit. However, many of these models and their learned representations are complex and therefore difficult for clinicians to interpret, creating challenges for validation. 
Our work proposes a new procedure to learn summaries of clinical timeseries that are both predictive and easily understood by humans. Specifically, our summaries consist of simple and intuitive functions of clinical data (e.g. ``falling mean arterial pressure'').  
Our learned summaries outperform traditional interpretable model classes and achieve performance comparable to state-of-the-art deep learning models on an in-hospital mortality classification task.  }

\section{Introduction}
Accurate predictions of patient risk in critical care units can aid clinicians in making more effective decisions.  Specifically, early identification of patients at high risk for in-hospital mortality is critical to assess patient disease acuity and inform life-saving interventions \cite{saps-ii, escobar2020}.  To predict in-hospital mortality risk, researchers have developed algorithms ranging from simple score-cards \cite{saps-ii, apache} to statistical machine learning (ML) models.  Recent advances in ML have led to the development of models with vast improvements in predictive accuracy for patient in-hospital mortality risk \cite{ awad2017early, gp-timeseries-acuity, ng-deep-learning-ehr, benchmarking-deep}.  

Despite these improvements, however, ML models are still prone to critical errors, often failing to generalize across different care settings or institutions \cite{futoma2020myth}.  An emerging line of research in interpretability, defined by  \cite{fdv-rigorous} as ``the ability to explain or to present in understandable terms to a human,'' provides an alternative way to ensure systems preserve properties such as safety or nondiscrimination.  If a model is interpretable to stakeholders, then clinical experts can inspect the model and verify that its reasoning is sound.  This ability to audit and validate is especially important when the models are used to inform critical decisions affecting patient health.  

In this work, we present a novel ML method to learn clinical timeseries summaries that are interpretable and predictive.  We introduce functions to compute summaries that align with simple and intuitive concepts, such as whether the timeseries is decreasing or spikes above a critical threshold. In contrast to prior work, our method learns how much of the timeseries should be used to calculate these summaries, discarding earlier timesteps that may be irrelevant for a specific prediction task.  Importantly, we introduce relaxations of our summary definitions to enable differentiable optimization, allowing summary parameters to be learned jointly with those of a downstream model.  
We show that with our method, we can achieve accuracies comparable with state-of-the-art baselines without sacrificing interpretability.

\section{Related Work}

Prior work on explaining clinical timeseries models fall into two categories: learning simple models that are inherently interpretable, and generating explanations of complex black box models.  We summarize a few key examples below.

\textbf{Interpreting Deep Models.} 
One popular strategy for explaining ML models is learning a second post-hoc explanation model to explain the first black box model \cite{ribeiro2016should}.  Many post-hoc explanation techniques for clinical timeseries models train explanation models that quantify the relative importance of each clinical variable \cite{NEURIPS2020_08fa4358, lundberg2018explainable}.  However, several works argue against the use of post-hoc explanation techniques, as explanation models are not always faithful or representative of the true underlying black boxes \cite{rudin2019stop, lakkaraju2020fooling}. 
Our method avoids these problems by design, instead explicitly optimizing for interpretability so that a second explanation model is not needed.

Another line of research proposes attention mechanism models specifically designed for timeseries.  \cite{choi2017retain, sha2017} present neural attention architectures for clinical timeseries and argue that attention scores measure feature importance. However, attention methods are often highly complex and nonlinear. Furthermore, \cite{serrano2019} shows attention scores do not always reflect true importance.  Instead of approximating importance, our study uses linear models over richer features, where importance does not need to be approximated but can be directly read off model coefficients.

\textbf{Expert Systems and Expert Features.}  Our work extends a long tradition of clinical experts hand-crafting features to create interpretable clinical decision-support algorithms.  Two expert systems widely used in ICUs are SAPS-II \cite{saps-ii} and APACHE \cite{apache}, which use simple score-card algorithms to evaluate patient acuity.  These systems use input features such as the patient's average or worst lab or vital values over time to compute mortality risk.  While SAPS-II and APACHE are simple and simulable to clinicians, their predictiveness is limited, as they cannot capture how labs or vitals change across a patient's stay.

A similar line of research proposes manual construction of expert features from clinical data, which are then used as input to ML models \cite{sun2012, roe2020}.  One limitation of this approach is that expert feature derivation is expensive and requires clinical expertise. Rather than relying on expert knowledge to identify which summary features will be the most predictive for a given task, our work instead uses optimization with a sparsity constraint to automatically learn which summary features are the most predictive.

\textbf{Summarizing Clinical Timeseries.}  A number of works have proposed a wide range of summary statistics for patient timeseries data.  Many works such as \cite{awad2017early, harutyunyan2019} train clinical models using the minimum value, maximum value, first measured value, or skew of clinical timeseries data.  \cite{DBLP:journals/midm/GuoLC20} proposes a more comprehensive set of 14 summary statistics to characterize the central tendency, dispersion tendency, and distribution shape of clinical timeseries data.  Our work extends this research, and is the first to our knowledge to use the slope of the timeseries or proportion of time above or below critical thresholds.  Our work is also novel in that we explicitly model and optimize for the duration over which we compute each summary feature.

\section{Cohort and Problem Set-Up}
Our goal is to learn summaries of patient timeseries data that are both human-interpretable and predictive for a downstream classification task.  Our approach will define how to calculate these human-interpretable summaries, and describe how both summary and classification model parameters are learned through optimization. In what follows, we detail each of these processes.

\textbf{Prediction Task.} Our work examines early prediction of in-hospital mortality.  We use the patient's first 24 hours of data to predict if they would later expire over the course of the remainder of their admission.  Patients who expired in the first 24 hours of their stay were excluded from our cohort.

\textbf{Cohort Selection.}
We use data from the MIMIC-III critical care database \cite{mimic-iii}, which contains deidentified health data from patients in critical care units of the Beth Israel Deaconess Medical Center between 2001 and 2012.  All data was extracted from MIMIC-III PhysioNet version 1.4, which contains 30,232 patients. We exclude patients under 18 years of age and patients whose weight is not measured.  We include data from each patient's first hospitalization only, and only patients with stays between 24-72 hours \cite{awad2017early}. After applying these criteria, our final cohort contained 11,035 patients, 15.23\% of whom died in-hospital. Cohort characteristics and demographics are summarized in Table \ref{table:mimic-summary-stats}.   
\begin{table}[h!]
\centering
\resizebox{\textwidth}{!}{%
\begin{tabular}{|c | c  c  | c  c  c  c | c  c  c  c|} 
 \hline
 Cohort & Age & \% Female & \% Urgent & \% Emergency & \% Elective & \% MICU & \% SICU & \% CCU & \% CSRU &   \\ 
 \hline
 All & 64.7 & 43.8 &  1.12 & 84.46 & 14.43 & 42 & 18 & 12 & 16 &\\\hline
 + & 70.9 & 46.5  & 0.77 & 96.25 & 2.97 & 53 & 18 & 13 & 4 &\\ \hline
 - & 64.1 & 43.6  & 1.14 & 83.22 & 15.64 & 41 & 18 & 12 & 17 & \\ \hline
\end{tabular}}
\caption{Mean statistics for the population cohort, and for cohorts of positive versus negative patients for in-hospital mortality.  Abbreviations: MICU, medical care unit; SICU, surgical care unit; CCU, cardiac care unit; CSRU, cardiac-surgery recovery
unit.}
\label{table:mimic-summary-stats}
\end{table}

\subsection{Extracting Inputs and Outputs}
\begin{figure}[]
\centering
    \includegraphics[width=0.6\textwidth]{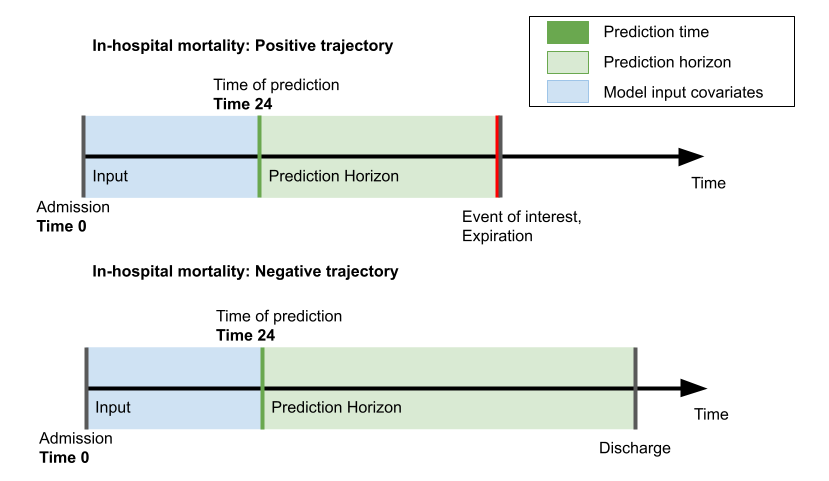}
\caption{Example positive and negative time-series to illustrate feature extraction.  The two trajectories have input data extracted from time 0 to time of prediction $T = 24$.  Figure inspired by Sherman et al \cite{sherman}.}
\label{figure:feature-extraction}
\end{figure}

For each patient $n$ in our $N$ patient cohort, we extracted static observations and physiological data including labs and vital signs sampled hourly. All clinical variables were separately normalized to have zero mean and unit variance.  Figure \ref{figure:feature-extraction} shows how features are extracted for patients that are positive versus negative for in-hospital mortality.  

\textbf{Static observations $\bm{S}$.} Matrix $\bm{S}: (N \times 8)$ contains $8$ demographic variables for each patient $n$: their age at admission, gender, and other information about their ICU stay (their first ICU service type, and whether their admission was urgent, emergency, or elective).

\textbf{Per-timestep clinical observations $\bm{X}$.}  The clinical variable tensor $\bm{X}: (N \times D \times T)$ contains $D = 28$ measurements of clinical variables for each of the $N$ patients at time $t$, discretized by hour.  These 28 measurements consist of vital signs and labs: diastolic blood pressure, fio2, GCS score, heartrate, mean arterial blood pressure, systolic blood pressure, SRR, oxygen saturation, body temperature,  urine output, blood urea nitrogen, magnesium, platelets, sodium, ALT and AST, hematocrit, po2, white blood cell count, bicarbonate, creatinine,  lactate, pco2, glucose, INR, hemaglobin, and bilirubin.  Missing values at timestep $t$ were imputed using either the most recent measurement of the variable, or the population median if the variable had not yet been measured during the patient's stay.  We use the patient's first $T = 24$ hours of data to predict in-hospital mortality.  We use subscripts to index into the tensor: for example, $\bm{X}_t$ indicates the $(N \times D)$ matrix of measurements taken at time $t$.


\textbf{Per-timestep measurement indicators $\bm{M}$.}
The measurement indicator tensor $\bm{M}: (N \times D \times T)$ contains indicator elements  $\bm{M}_{n, d, t}$ which are 1 if their corresponding clinical variable in $\bm{X}_{n, d, t}$ was measured at time $t$, 0 otherwise.

\textbf{Outcome labels $\bm{y}$.} Label vector $\bm{y}$ contains indicators $\bm{y}_n$ which are 1 if patient $n$ expired in-hospital, 0 otherwise.

\section{Methods}
\begin{figure}[]
\centering
    \includegraphics[width=0.7\textwidth]{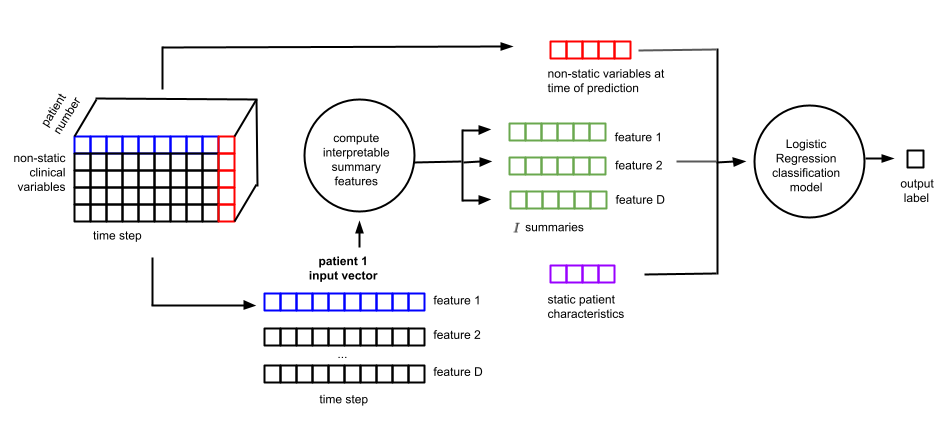}
\caption{Illustrates summary extraction for prediction from timeseries data.  First, non-static clinical variables $\{\bm{X}, \bm{M}\} $ are used to compute interpretable summary features $\bm{H}$.  Then, summary features $\bm{H}$, static features $\bm{S}$, and the non-static variables $\{\bm{X}, \bm{M}\}$ at the time of prediction are given as input to a Logistic Regression model $g$, which predicts output labels $\bm{\hat{y}}$.  Figure inspired by Ghassemi, Szolovits et al \cite{hughes-vaso}.}
\label{figure:model-diagram}
\end{figure}

Given a cohort of $N$ training examples $\{ \bm{X}, \bm{M}, \bm{S}, \bm{y} \}$, we propose a novel procedure for learning predictive human-interpretable timeseries summaries. Concretely, we first compute summaries $\bm{H}$ from clinical timeseries data $(\bm{X}, \bm{M})$. We then use the summaries $\bm{H}$ in addition to static data $\bm{S}$ and clinical variables $\bm{X}$ as input to a Logistic Regression model to predict labels $\bm{y}$.  This process of using human-interpretable summaries for prediction is shown in Figure \ref{figure:model-diagram}.

In Section \ref{section:model-def}, we motivate and introduce our novel summary features.  In Section \ref{section:relaxations}, we give continuous relaxations of summary feature definitions to enable efficient inference of summary parameters.  In Section \ref{section:learning-process}, we discuss how predictive summary features can be jointly learned with downstream classification model parameters.

\subsection{Model: Defining human-interpretable summaries}\label{section:model-def}
Measures of central tendency and dispersion have commonly been used to summarize timeseries (see survey in \cite{DBLP:journals/midm/GuoLC20}).  Our key modelling innovations include adding additional features that correspond to how clinicians themselves describe timeseries.  Our novel contributions include explicitly modelling the overall trend of a lab/vital and the number of hours that a lab/vital dips above or below a threshold, as well as allowing different features to be computed over different periods of time (e.g. the most recent 6 hours vs. the most recent 24 hours).

The $I = 13$ summary statistics used in this study are listed in Table \ref{table:summaries-weighted}.  Each of the summary statistics takes into account measurement indicators $\bm{M}$ so that clinical variable summaries are computed only using timesteps where the variable is measured. Each summary statistic is applied to each of the $D$ clinical variables to create the summary feature tensor $\bm{H}: (N \times D \times I)$.

Below, we expand on the parameterization of our summary features.  Next, we describe how we enable efficient, automated search over summary feature parameters.  Importantly, our approach automates many processes associated with summary design, enabling optimization over summary parameters. 

\textbf{Incorporating Duration.}
Many prior works in clinical timeseries modelling do not use all timesteps for the patient, but instead only the most recent available data, such as the six or twelve hours before the time of prediction.  In contrast to prior works, we explicitly model how much of each timeseries  should be used to compute each of the $I$ summaries in $\bm{H}$.  For example, we may wish to exclude earlier measurements of a particular clinical variable if only the variable's recent measurements before time of prediction $T$ are significant for a prediction task.  Specifically, for each variable $d$ and for each summary function $i$, we define a duration time $\bm{C}_{d, i}$.  Only the variable's timeseries data that occurred in the immediately previous $\bm{C}_{d, i}$ hours before the time of prediction is used to calculate summary $i$.  We organize all the duration time parameters $\bm{C}_{d,i}$ into a $(D \times I)$ matrix $\bm{C}$.

To exclude data that occurs before time $(T -\bm{C}_{d, i})$ when computing summary features, we multiply each of the original timeseries variables by indicator variables for whether the measurements occurred within $\bm{C}_{d, i}$ hours before time of prediction $T$.  For example, a mean summary statistic would be computed using indicator variables $\mathbbm{1}(\cdot)$ as:
\begin{equation}\label{non_diff_cutoff_indicators} 
\bm{H}_{i = mean}: (N \times D) = \left( \textstyle\sum_{t=1}^{T}  \mathbbm{1}(t > T - \bm{C}_{i = mean}) \odot \bm{X}_{ t} \odot \bm{M}_{ t} \right) / \left( \textstyle\sum_{t = 1}^{T} \bm{M}_{ t} \odot \mathbbm{1}(t > T - \bm{C}_{i = mean})  \right)
\end{equation}

where $\odot$ is the element-wise multiplication operator and division is performed element-wise.  Our objective is to learn duration parameters $\bm{C}$ that maximize the predictiveness of their corresponding summary features $\bm{H}$.

\textbf{Threshold Parameters.}
Some of the summary functions $f_i$ in Table \ref{table:summaries-weighted} have additional parameters such as thresholds.  For example, one of the summaries is the proportion of the patient's measured timeseries where their measured clinical variables are above some $D$-dimensional critical threshold parameter vector $\bm{\phi}^{+}$ for each variable:
\begin{equation}\label{thresholds-non-diff}
    \left( \textstyle\sum_{t = 1}^{T}  \bm{M}_t \odot \mathbbm{1} (\bm{X}_t > \bm{\phi}^{+}) \right) / \left( \textstyle\sum_{t = 1}^{T}  \bm{M}_t \right)
\end{equation}
These summaries correspond to clinically-intuitive ideas, such as whether the patient been mostly well or sick.  As with durations, we learn threshold parameters automatically to avoid burdening experts and to assist in prediction.

\subsection{Continuous Relaxations for Efficient Inference}\label{section:relaxations}
\textbf{Learning Duration Parameters.}
In Equation \ref{non_diff_cutoff_indicators}, we showed how summary functions $f_i$ that only use the most recent $\bm{C}$ hours of data can be calculated using indicator variables $\mathbbm{1}(t > T - \bm{C}_{d, i}) $.  These indicator variables, however, do not have informative gradients and are not differentiable.  To enable differentiable optimization for duration time parameters $\bm{C}$, we introduce weight parameters $\bm{W}$ by relaxing the indicator random variables using the sigmoid function $\sigma(x) = \frac{1}{1+e^{-x}}$.  Using the duration parameter matrix $\bm{C}$, we define $D$-dimensional vectors $\bm{w}_{t, i}$ that compose weight tensor $\bm{W}: (T \times I \times D)$: 

\begin{equation}\label{w_def}
    \bm{w}_{t, i}  = \sigma \left((t - T + \bm{C_i})/ \tau \right) 
\end{equation}

For each feature $d$ and summary $i$, clinical observations $\bm{X}_{ d, t}$ where $t > T - \bm{C}_{d, i}$ will have corresponding weights $\bm{w}_{t, i, d}$ near 1.  Timesteps $t$ where $t < T - \bm{C}_{d, i}$ will have corresponding weights $\bm{w}_{t, i, d}$ near 0.  Temperature parameter $\tau$ controls the harshness of the weight matrix: small temperatures push the sigmoid function towards its edges, learning weights that are closer to exactly 0 for timesteps before $T -\bm{C}_{d, i}$ and 1 for timesteps after $T - \bm{C}_{d, i}$, effectively functioning as the indicator variables in Equation \ref{non_diff_cutoff_indicators}.

Weighted summary functions $f_i$ used to derive human-interpretable summaries $\bm{H}$ can be found in Table \ref{table:summaries-weighted}. Duration parameters $\bm{C}$ that determine weight tensor $\bm{W}$ are included in $\beta_{\bm{H}}$, the set of all parameters necessary to compute the summaries.
\begin{table}[h!]
\centering

\begin{tabular}{|c| c|} 
 \hline
 Description & Function \\ [0.5ex] 
 \hline
 
 Mean of the time-series &   $\left(\sum_{t=1}^{T} \left( \bm{w}_{t, i} \odot \bm{X}_t \odot \bm{M}_t \right)\right) /  \left(\sum_{t = 1}^{T} \bm{M}_t \odot \bm{w}_{t, i} \right) $\\ \hline
 
 Variance of the time-series &   $ \frac{(\sum_{t = 1}^{T} \bm{M}_t \odot \bm{w}_{t, i})^2}{(\sum_{t=1}^{T} \bm{M}_t \odot \bm{w}_{t, i})^2 - \sum_{t=1}^{T} \bm{M}_t \odot \bm{W}^2_t}   \odot \sum_{t = 1}^{T} \bm{M}_t \odot \bm{w}_{t, i} \odot  \left( \bm{X}_t {-} \bar{\bm{X}} \right)^2$ \\ \hline
 
 
 Indicator if feature was ever measured &  $\sigma \left( \left( \sum_{t = 1}^T \bm{w}_{t, i} \odot \bm{M}_t \right) / \left(\tau \odot \sum_{t = 1}^T \bm{w}_{t, i}  \right)\right)$\\ \hline
 
 
 Mean of the indicator sequence &    $ \left(\sum_{t = 1}^{T} \bm{w}_{t, i} \odot \bm{M}_t \right) / \left(\sum_{t = 1}^{T} \bm{w}_{t, i} \right)$\\ \hline
 
Variance of the indicator sequence &  $ \left( \frac{(\sum_{t=1}^{T} \bm{w}_{t, i})^2}{(\sum_{t=1}^{T} w_t)^2 - \sum_{t=1}^{T} \bm{W}^2_t} \right)  \sum_{t = 1}^{T} \bm{w}_{t, i} \odot ( \bm{M}_t - \bar{\bm{M}})^2$\\ \hline
 
 \# switches from missing to measured &  $\left(\sum_{t = 1}^{T - 1} \bm{w}_{t, i} \odot  | \bm{M}_{t + 1} - \bm{M}_t|\right) / \left(\sum_{t=1}^{T} \bm{w}_{t, i} \right) $\\ \hline
 
 First time the feature is measured &  $ \min t$ s.t. $\bm{M}_t = 1$\\ \hline
 
 Last time the feature is measured &  $ \max t$ s.t. $\bm{M}_t = 1$\\ \hline
 
 Proportion of time above threshold $\bm{\phi}^{+}$ &  $  \left(  \sum_{t = 1}^{T} \bm{w}_{t, i} \odot \bm{M}_t \odot \sigma( \frac{\bm{X}_t- \bm{\phi}^{+}}{\tau}) \right) / \left( \sum_{t = 1}^{T} \bm{M}_t \odot \bm{w}_{t, i} \right)$ \\ \hline
 
 Proportion of time below threshold $\phi^{-}$ &  $   \left( \sum_{t = 1}^{T} \bm{w}_{t, i} \odot \bm{M}_t \odot \sigma( \frac{\bm{\phi}^{-} - \bm{X}_t }{\tau}) \right) / \left( \sum_{t = 1}^{T} \bm{M}_t \odot \bm{w}_{t, i} \right)$\\ \hline
 
 Slope of a L2 line & $\frac{\sum_{t= 1}^T \bm{w}_{t, i} (t - \bar{t}_{\bm{w}}) (\bm{X}_t - \bar{\bm{X}}_{\bm{w}})}{\sum_{t = 1}^T \bm{w}_{t, i} (t - \bar{t}_{\bm{w}})^2}$, where $\bar{t}_{\bm{w}} = \frac{\sum_t \bm{w}_{t, i} \cdot t}{\sum_t \bm{w}_{t, i}}$ and $\bar{\bm{X}}_{\bm{w}} = \frac{\sum_t \bm{w}_{t, i} \odot \bm{X}_t}{\sum_t \bm{w}_{t, i}}$ \\ \hline
 
 
 Standard error of the L2 line slope &   $1 / \left( \sum_t \bm{w}_{t, i} \odot (t - \bar{t}_{\bm{w}})^2 \right)$  \\ \hline
 
 
 

 \hline
\end{tabular}
\caption{Table of functions $f_i$ used to calculate human-interpretable summaries $\bm{H}$.  For each of the $D$ clinical variables, all $I$ of the above functions are applied to each of the $N$ patients.  Each of the $I$ summary features $i$ is defined with respect to $D$-dimensional weight vectors $\bm{w}_{t, i}$ defined in Section \ref{section:relaxations}.  Paramater $\tau$ is a temperature parameter for the sigmoid function.   $\mathbbm{1}(\cdot)$ denotes indicator variables for events inside the parentheses, $\odot$ indicates element-wise matrix multiplication and division is done element-wise.  Additionally, $\bar{\bm{X}} = \sum_t^T \bm{M}_t \odot \bm{X}_t /\sum_t^T \bm{M}_t$, and  $\bar{\bm{M}}_t = \frac{1}{T} \sum_t^T \bm{M}_t$.}
\label{table:summaries-weighted}
\end{table}

 \textbf{Learning Threshold Parameters.}
Our work relaxes summary definitions to enable differentiable optimization to learn summary parameters.
The indicator variables used to define the proportion of hours that a patient's timeseries is above thresholds $\bm{\phi}^{+}$ in Equation \ref{thresholds-non-diff} are non-differentiable with respect to $\bm{\phi^{+}}$. To enable differentiable optimization, our work defines our threshold summary features using the sigmoid function $\sigma$ with temperature parameter $\tau$:
\begin{equation}
    f_{threshold}(\bm{X}, \bm{M}, \bm{W}) =   \left(\textstyle\sum_{t = 1}^{T} \bm{w}_{t, i = threshold} \odot \bm{M}_t \odot \sigma\left( \frac{\bm{X}_t - \bm{\phi}^{+}}{\tau}\right)\right) / \left( \textstyle\sum_{t = 1}^{T} \bm{M}_t \cdot \bm{w}_{t, i = threshold} \right) 
\end{equation}

Threshold parameters $\{\bm{\phi}^{+}, \bm{\phi}^{-}\}$ are included in $\beta_{\bm{H}}$, the set of all parameters necessary to compute the summaries.

\subsection{Learning Process}\label{section:learning-process}
Our study uses summary features $\bm{H}$, along with static variables $\bm{S}$ and timeseries variables $\{\bm{X}, \bm{M}\}$ at prediction time $T = 24$ as input to a Logistic Regression model $g$ with coefficients $\bm{\beta}_g$. Logistic Regression model $g$ outputs predicted probabilities $\hat{\bm{y}} = p(\bm{y} = \bm{1} | \bm{X},\bm{S}, \bm{M})$.  Our objective is to learn optimal summary and model parameters $\bm{\beta} = \{\bm{\beta}_{\bm{H}}, \bm{\beta}_g\}$.  We jointly learn parameters $\bm{\beta}$ by minimizing the loss function
\begin{equation}
    \mathcal{L}(\bm{\beta}; \bm{X}, \bm{M}, \bm{S}, \bm{y}) = - \frac{1}{N} \sum_{n = 1}^N \omega_n \left( \bm{y}_n \cdot \log[g(\bm{X}_n, \bm{M}_n, \bm{S}_n, \bm{\beta})] + (1 - \bm{y}_n) \log[1 - g(\bm{X}_n, \bm{M}_n, \bm{S}_n, \bm{\beta})] \right)  + \Omega(\bm{\beta}_g)
\end{equation} 
Our loss function is the sum of the weighted binary cross-entropy loss using predictive model $g$ and the regularization penalty $\Omega(\bm{\beta}_g)$.  To account for class imbalance, we reweight each training example $n$'s loss contribution by $\omega_n$, the inverse of its class frequency in the training dataset.

\textbf{Horseshoe Regularization on Coefficients.}
We explicitly optimize for sparsity in model parameters $\bm{\beta}_g$ via our regularization penalty in our training objective.  We use a Horseshoe regularization penalty $\Omega(\bm{\beta}_g)$ with shrinkage parameter $1$ to encourage sparsity in the learned regression coefficients\cite{horseshoe-bhadra}.    

\section{Experiments}
We compare models learned using our summaries to other interpretable models as well as deep learning baselines. We show that our models outperform other traditional human-interpretable model classes and achieve performance comparable to deep models on the in-hospital mortality task.

\textbf{Model configurations.} Our baseline models are: Ridge Logistic Regression models that take as input only the patient timeseries measured at the time of prediction $T$, and Ridge Logistic Regression and LSTM models that take as input all of the patient timeseries.  We used an LSTM as our deep baseline architecture as prior works document their superior performance at mortality prediction from clinical timeseries data \cite{Harutyunyan19}.

Our LSTM models were trained on the sequential timeseries data $\{\bm{X}, \bm{M}\}$ with a step size of 1 hour.  The LSTM hidden states at each timestep were then used to predict both the next timestep of the patient timeseries $\bm{X}$, and to predict outcome labels $\bm{y}$.  All $T$ of the output hidden states $\bm{X}_t$ were input to a fully-connected layer, which output predictions of the next timestep $\bm{X}_{t + 1}$.  The last output hidden state at time $T$ was also input with static features $\bm{S}$ to a fully-connected layer to predict outcome labels $\bm{y}$.  The LSTM models were trained to minimize both the mean-squared error of the next state prediction, and the binary cross entropy loss of the classification prediction.  ReLU activation functions were applied to both fully-connected layers.

\textbf{Training Details.}
For training and testing, we split the cohort of $N$ patients into train and test sets, where all data associated with each patient is either in train or test.  All performance metrics are averaged across five train-test splits.  Ridge baseline models were implemented using RidgeCV from \texttt{scikit-learn} \cite{scikit-learn}. LSTMs as well as our summary-based Logistic Regression models were implemented with PyTorch, and trained with the Adam optimizer\cite{adam} at a batch size of 256.  We trained all of our Logistic Regression models for 30,000 epochs and LSTM models for 10,000 epochs using early stopping.  All hyperparameters (including the LSTM hidden state and layer dimensions, optimizer learning rate, and regularization parameters) were selected via random hyperparameter search \cite{bergstra2012random}.  All temperature parameters $\tau$ were set to $0.1$. 

Our final learned LSTM models have hidden state dimension 32, 1024 nodes in the layer to predict the next timestep, and 64 nodes in the layer to predict labels $\bm{y}$.  They are trained using a learning rate of $1e-05$.  Our final learned models with summaries use $\alpha = 1e-05$, Horseshoe shrinkage parameter $1.0$, and learning rate $1e-05$.  

\section{Results}
\textbf{Our learned models achieve performance comparable to state-of-the-art baselines.} Table \ref{table:mortality-auc} compares the performance of our learned models with linear and deep baselines for the in-hospital mortality prediction task. Applying a linear model to the learned summary features $\bm{H}$ consistently improves AUCs in comparison to using a linear model on clinical timeseries and static data alone. Our models achieve an AUC performance comparable to state-of-the-art LSTM models with test AUCs of $0.9000 \pm 0.0223$. Notably, our models that allow differentiable optimization to learn duration times outperform models that compute all summary features using the entire duration of the patient timeseries.  This implies that there is predictive value in explicitly modelling how much of each variable's clinical timeseries should be considered for a specific prediction task.
\begin{table}[h!]
\centering
\begin{tabular}{|c | c| c |} 
 \hline
 \textbf{Model} & Train set AUC & Test set AUC\\ \hline
 LR trained on $\{\bm{S}, \bm{M}\}$ and $\bm{X}$ at time $T$ only & $0.8653 \pm 0.0013$ & $0.8626 \pm 0.0079$ \\ \hline
 LR trained on $\{\bm{S}, \bm{M}, \bm{X}\}$ & $0.8931 \pm 0.0015$	& $ 0.8668 \pm 0.0122$ \\ \hline
 LSTM trained on $\{\bm{S}, \bm{M}, \bm{X}\}$ & $\bf{0.9101 \pm 0.0018}$ & $\bf{0.9000 \pm 0.0223}$ \\ \hline
 Our model, trained on $\{\bm{S}, \bm{M}, \bm{X}\}$, non-differentiable durations $\bm{C}$ & $\bf{0.9065 \pm 0.0019}$ & $\bf{0.8818 \pm 0.0063}$ \\ \hline 
  Our model, trained on $\{\bm{S}, \bm{M}, \bm{X}\}$, differentiable durations $\bm{C}$ & $\bf{0.9074 \pm 0.0016}$ & $\bf{ 0.8867 \pm 0.0061}$ \\ \hline
\end{tabular}

\caption{Performance of learned models on the in-hospital mortality prediction task.  AUCs are averaged over five train-test splits with their standard error.  Abbreviations: LR, Logistic Regression; LSTM, long short-term memory. $\bm{C}$ refers to the duration parameters defined in Section \ref{section:relaxations}.}
 
\label{table:mortality-auc}
\end{table}

\begin{figure}[]
\centering
    \includegraphics[width=0.5\textwidth]{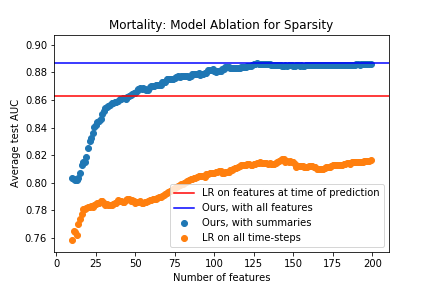}
\caption{Prediction quality (mean test AUC) vs. model complexity (number of non-zero features) for baseline Ridge Regression trained on all timesteps of the patient timeseries, versus our models.  The dark blue horizontal line shows our model's average test set AUC using all $401$ derived features, and the orange horizontal line shows the average test set AUC of the Logistic Regression model trained only on the $65$ features extracted at the time of prediction.}
\label{figure:ablation-sparsity}
\end{figure}

\textbf{Our learned models use fewer features to achieve higher accuracy in comparison to other interpretable baselines.} To evaluate the sparsity of each model, we performed a set of ablation experiments where we zeroed all but the $N$ coefficients with the largest magnitudes for each of the learned Logistic Regression models.  In Figure \ref{figure:ablation-sparsity}, we show the average test set AUC for Logistic Regression baseline versus our models using only $N$ coefficients.  Our models consistently outperform baseline models when all but $N$ coefficients are zeroed, suggesting that our models learn a smaller and more predictive set of important features.  

\textbf{Our learned models are interpretable.} 
Table \ref{table:key-summary-features} shows 15 key summary features that consistently have the largest learned Logistic Regression coefficients across train-test splits. The corresponding coefficients for each feature can be interpreted as a measure of the feature's contribution to the final classification label.  For example, because the mean of the patient's GCS has a large negative coefficient, this means that patients with higher mean GCS scores will be assigned lower predicted probabilities for in-hospital mortality.  Therefore our models are \emph{decomposable} \cite{lipton-mythos}, as each of the model's features and coefficients has an intuitive clinical explanation.


\textbf{Our learned summary features are clinically sensible.} 
The vast majority of the key summary features learned by our models shown in Table \ref{table:key-summary-features} are supported by studies in medical literature. For instance, it is widely accepted that patients who are older tend to have lower chances of survival in ICU settings  \cite{nielsen2019survival,fuchs2012icu}. Similarly, patients with lower GCS scores of below 6 tend to have severe injuries and higher chances of mortality \cite{bastos1993glasgow}. Notably a lower GCS score in the later hours of a patient's hospitalisation significantly reduces a patient's chances of survival \cite{10.1001/archneur.1990.00530110035013,settervall2011hospital}. Finally, the normal range of features such as the blood oxygen saturation (SPO$_2$) is between 95\% and 100\%. An SPO$_2$ consistently below 90\% indicates hypoxaemia or potential respiratory distress. These patients have to be mechanically ventilated in ICU and frequently have lower chances of survival \cite{lazzerini2015hypoxaemia,vold2015low}.
\begin{table}[h!]
\centering
\begin{tabular}{|c | l| l | r | } 
 \hline
 \textbf{Feature} & \textbf{Aggregation} & \textbf{Time} & \textbf{Coefficient} \\ \hline
 Age & static value & - &  14.9 \\ \hline
 BUN & value at & hour 24 & 6.2 \\ \hline
 GCS & mean over & hours 5 - 24 & - 5.59 \\ \hline
 HR & value at & hour 24 & 4.69 \\ \hline
 FiO$_2$ & value at & hour 24 & 4.31 \\\hline
 Hct & times measured over & hours 2 - 24 & - 3.81 \\  \hline
 HR, & mean over & hours 2 - 24 & 3.78 \\ \hline
 GCS & value at & hour 24 & - 3.62 \\ \hline
 GCS & hours below 6.08 & hours 7 - 24 & 3.37 \\\hline
 Creatinine & hours below 0.35 mg/dL  & hours 5 - 24 & 2.76 \\ \hline
 FiO$_2$ & hours above 62.96\% & hours 16 - 24 & 2.59 \\ \hline
 SpontaneousRR & mean over & hours 2 - 24 & 2.29 \\ \hline
 GCS & times measured over & hours 5 - 24 & 2.23 \\ \hline
 Sodium & hours below 131.57 mEq/L & hours 1 - 24 & 2.16 \\\hline
 WBC & hours below 0.78 cells/mL &  hours 6 - 24 & 2.10 \\\hline
 SPO$_2$ & hours below 92.36\% & hours 10 - 24 & 2.04 \\  \hline
\end{tabular}

\caption{ Key summary features, sorted from largest to smallest coefficient magnitudes, from learned models.  }
 
\label{table:key-summary-features}
\end{table}

\textbf{Initialization sensitivity.}
In general, we observed that our optimization procedure is stable, learning the same 15 key summary features across different stochastic parameter initializations and train-test splits.  However, there are cases where we observed that the learned duration parameters $\bm{C}$ varied depending on their initialization. As such, we recommend that practitioners incorporate prior knowledge about the clinical prediction task when initializing the duration time parameters.  For example, if examining the entire duration of the patient's timeseries is necessary for a prediction task, then the duration parameters should be initialized to include the entire timeseries by default.  

\section{Discussion 
\& Conclusion}
In this work, we defined functions to compute interpretable, parameterizable summaries of clinical timeseries, and developed relaxations so that our summary parameters could be jointly learned with a downstream predictive model.  In our experiments, we used Logistic Regression to make predictions because its coefficients are easily decomposable \cite{lipton-mythos}.   However, because our learned summaries are inherently interpretable, any other interpretable architecture could be used instead.  Our methodology is generalizable and enables the efficient learning of intuitive and predictive timeseries summaries without placing any assumptions on the downstream model architecture.

\textbf{Future work.} Our study poses many interesting directions for future work.  One avenue would be to conduct a user study to validate the human-interpretability and decomposability of our proposed summary features.  Another would be to evaluate whether the summary features learned for particular critical care prediction tasks remain predictive for a wider set of critical care prediction tasks. 
Finally, we could also develop additional summary statistic functions, or expand our framework to consider sharing duration parameters across features or across summaries to better model dependencies between clinical labs and vitals---as many physiological events are characterized by several simultaneous changes to multiple labs and vitals \cite{hotchkiss2017septic}.

\textbf{Conclusion.}  In this paper, we propose a new method to learn interpretable and predictive summary features from clinical timeseries data.  In addition to introducing novel summary statistics including slope and threshold features, our work differs from prior work by learning the duration of timeseries data that should be used to compute each summary.  We demonstrate that our learned timeseries summaries achieve performance quality comparable to state-of-the-art deep models when trained to predict early patient mortality risk on real patient data.  We also qualitatively validate our models to confirm their interpretability and sensibility. Our work is an important step towards optimizing for  representations of clinical timeseries data that are both highly predictive and interpretable.

\textbf{Acknowledgements:} NJ and FDV acknowledge support from NIH R01 MH123804-01A1. SP acknowledges support from the Miami Foundation and SNSF P2BSP2-184359.

\bibliographystyle{unsrt}

\end{document}